\documentclass[runningheads]{llncs}
\usepackage{amsfonts}
\usepackage{graphicx}
\usepackage{amsfonts}
\usepackage{graphicx}
\usepackage{epsfig}
\usepackage{graphicx}
\usepackage{amsmath}
\usepackage{amssymb}
\usepackage{mathtools}
\usepackage{resizegather}
\usepackage{marvosym}
\usepackage[linesnumbered,ruled]{algorithm2e}
\usepackage{dsfont}
\usepackage{enumitem}
\usepackage{caption}
\usepackage{multirow}
\RequirePackage[utf8]{inputenc}

\usepackage{footnote}
\usepackage[hang, flushmargin]{footmisc}
\usepackage{footnotebackref}
\begin{document}
\title{Adapting Off-the-Shelf Source Segmenter for Target Medical Image Segmentation}
\titlerunning{Batch statistics for Source-Relaxed Segmentation UDA}
 
\author{Xiaofeng Liu\inst{1} \and Fangxu Xing\inst{1} \and Chao Yang\inst{2} \and Georges El Fakhri\inst{1}  \and~~~ Jonghye Woo\inst{1}}

%index{Xiaofeng, Liu}
%index{Fangxu, Xing}
%index{Chao, Yang}
%index{Georges, El Fakhri}
%index{Jonghye, Woo}

\institute{Gordon Center for Medical Imaging, Department of Radiology, Massachusetts General Hospital and Harvard Medical School, Boston, MA, 02114\and
Facebook Artificial Intelligence, Boston, MA, 02142}

\authorrunning{X. Liu et al.}

\maketitle              % typeset the header of the contribution

\begin{abstract}

Unsupervised domain adaptation (UDA) aims to transfer knowledge learned from a labeled source domain to an unlabeled and unseen target domain, which is usually trained on data from both domains. Access to the source domain data at the adaptation stage, however, is often limited, due to data storage or privacy issues. To alleviate this, in this work, we target source free UDA for segmentation, and propose to adapt an ``off-the-shelf" segmentation model pre-trained in the source domain to the target domain, with an adaptive batch-wise normalization statistics adaptation framework. Specifically, the domain-specific low-order batch statistics, i.e., mean and variance, are gradually adapted with an exponential momentum decay scheme, while the consistency of domain shareable high-order batch statistics, i.e., scaling and shifting parameters, is explicitly enforced by our optimization objective. The transferability of each channel is adaptively measured first from which to balance the contribution of each channel. Moreover, the proposed source free UDA framework is orthogonal to unsupervised learning methods, e.g., self-entropy minimization, which can thus be simply added on top of our framework. Extensive experiments on the BraTS 2018 database show that our source free UDA framework outperformed existing source-relaxed UDA methods for the cross-subtype UDA segmentation task and yielded comparable results for the cross-modality UDA segmentation task, compared with a supervised UDA methods with the source data.
\end{abstract}

\section{Introduction} 

Accurate tumor segmentation is a critical step for early tumor detection and intervention, and has been significantly improved with advanced deep neural networks (DNN) \cite{wang2021automated,liu2018ordinal,liu2019unimodal,liu2020unimodal,liu2021symmetric}. A segmentation model trained in a source domain, however, usually cannot generalize well in a target domain, e.g., data acquired from a new scanner or different clinical center, in implementation. Besides, annotating data in the new target domain is costly and even infeasible \cite{liu2021Generalization}. To address this, unsupervised domain adaptation (UDA) was proposed to transfer knowledge from a labeled source domain to unlabeled target domains \cite{liu2021subtype}. 

The typical UDA solutions can be classified into three categories: statistic moment matching, feature/pixel-level adversarial learning \cite{liu2021dual,liu2021unified,liu2020energy}, and self-training \cite{zou2019confidence,liu2021Self-training}. These UDA methods assume that the source domain data are available and usually trained together with target data. The source data, however, are often inaccessible, due to data storage or privacy issues, for cross-clinical center implementation \cite{bateson2020source}. Therefore, it is of great importance to apply an ``off-the-shelf" source domain model, without access to the source data. For source-free classification UDA, Liang et al.~\cite{liang2020we} proposed to enforce the diverse predictions, while the diversity of neighboring pixels is not suited for the segmentation purpose. In addition, the class prototype \cite{liu2021subtype} and variational inference methods~\cite{liu2021Generalization} are not scalable for pixel-wise classification based segmentation. More importantly, without distribution alignment, these methods relied on unreliable noisy pseudo labeling. 

Recently, the source relaxed UDA \cite{bateson2020source} was presented to pre-train an additional class ratio predictor in the source domain, by assuming that the class ratio, i.e., pixel proportion in segmentation, is invariant between source and target domains. At the adaptation stage, the class ratio was used as the only transferable knowledge. However, that work \cite{bateson2020source} has two limitations. First, the class ratio can be different between the two domains, due to label shift \cite{liu2021Generalization,liu2021subtype}. For example, a disease incident rate could vary between different countries, and tumor size could vary between different subtypes and populations. Second, the pre-trained class ratio predictor used in \cite{bateson2020source} is not typical for medical image segmentation, thereby requiring an additional training step using the data in the source domain. 

In this work, to address the aforementioned limitations, we propose a practical UDA framework aimed at the source-free UDA for segmentation, without an additional network trained in the source domain or the unrealistic assumption of class ratio consistency between source and target domains. More specifically, our framework hinges on the batch-wise normalization statistics, which are easy to access and compute. Batch Normalization (BN) \cite{ioffe2015batch} has been a default setting in the most of modern DNNs, e.g., ResNet \cite{He_2016_CVPR} and U-Net \cite{zhou2019normalization}, for faster and more stable training. Notably, the BN statistics of the source domain are stored in the model itself. The low-order batch statistics, e.g., mean and variance, are domain-specific, due to the discrepancy of input data. To gradually adapt the low-order batch statistics from the source domain to the target domain, we develop a momentum-based progression scheme, where the momentum follows an exponential decay w.r.t. the adaptation iteration. For the domain shareable high-order batch statistics, e.g., scaling and shifting parameters, a high-order batch statistics consistent loss is applied to explicitly enforce the discrepancy minimization. The transferability of each channel is adaptively measured first, from which to balance the contribution of each channel. Moreover, the proposed unsupervised self-entropy minimization can be simply added on top of our framework to boost the performance further. 

Our contributions are summarized as follows:

\noindent$\bullet$ To our knowledge, this is the first source relaxed or source free UDA framework for segmentation. We do not need an additional source domain network, or the unrealistic assumption of the class ratio consistency \cite{bateson2020source}. Our method only relies on an ``off-the-shelf" pre-trained segmentation model with BN in the source domain.

%We propose a practical segmentation UDA framework in the absence of source domain data, which 

\noindent$\bullet$ The domain-specific and shareable batch-wise statistics are explored via the low-order statistics progression with an exponential momentum decay scheme and transferability adaptive high-order statistics consistency loss, respectively. %The unsupervised self-entropy minimization can be simply added. 

\noindent$\bullet$ Comprehensive evaluations on both cross-subtype (i.e., HGG to LGG) and cross-modality (i.e., T2 to T1/T1ce/FLAIR) UDA tasks using the BraTS 2018 database demonstrate the validity of our proposed framework and its superiority to conventional source-relaxed/source-based UDA methods.

\begin{figure}[t]
\begin{center}
\includegraphics[width=1\linewidth]{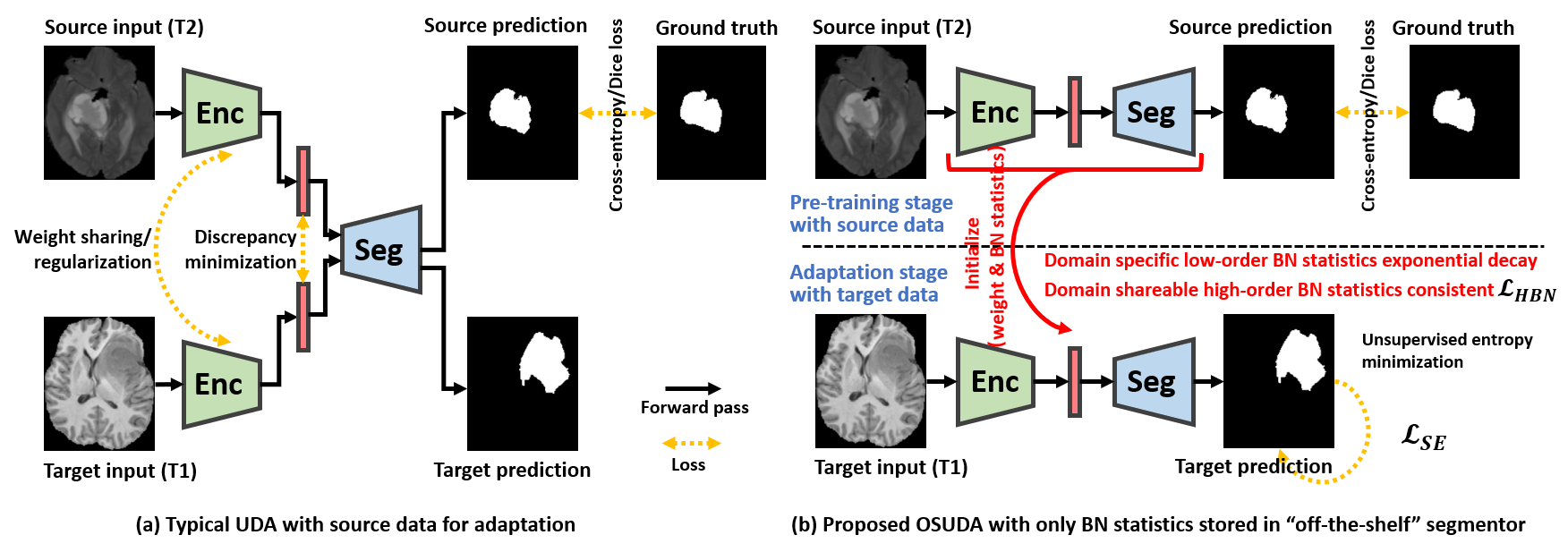}
\end{center}  
\caption{Comparison of (a) conventional UDA \cite{wilson2020survey} and (b) our source-relaxed OSUDA segmentation framework based on the pre-trained ``off-the-shelf" model with BN. We minimize the domain discrepancy based on the adaptively computed batch-wise statistics in each channel. The model consists of a feature encoder (Enc) and a segmentor (Seg) akin to \cite{chen2019synergistic,zou2020unsupervised}.
} 
\label{fig1}\end{figure}

\section{Methodology}

We assume that a segmentation model with BN is pre-trained with source domain data, and the batch statistics are inherently stored in the model itself. At the adaptation stage, we fine-tune the model based on the batch-wise statistics and the self-entropy (SE) of target data prediction. The overview of the different setups of conventional UDA and our ``off-the-shelf (OS)" UDA is shown in Fig. \ref{fig1}.
Below, we briefly revisit the BN in Subsec. 2.1 first and then introduce our OSUDA in Subsec. 2.2. The added unsupervised SE minimization and the overall training protocol are detailed in Subsec. 2.3.
  
\subsection{Preliminaries on Batch Normalization} 

As a default setting in the most of modern DNNs, e.g., ResNet \cite{He_2016_CVPR} and U-Net \cite{zhou2019normalization}, Batch Normalization (BN) \cite{ioffe2015batch} normalizes the input feature in the $l$-th layer $f_l\in\mathbb{R}^{B\times H_l\times W_l\times C_l}$ within a batch in a channel-wise manner to have zero mean and unit variance. $B$ denotes the number of images in a batch, and $H_l, W_l,$ and $C_l$ are the height, width, and channels of layer $l$. We have samples in a batch, with index $b\in\{1,\cdots,B\}$, spatial index $n\in\{1,\cdots,H_l\times W_l\}$, and channel index $c\in\{1,\cdots,C_l\}$. BN calculates the mean of each channel $\mu_{l,c}=\frac{1}{B\times H_l\times W_l}\sum_b^B\sum_n^{H_l\times W_l}f_{l,b,n,c}$, where $f_{l,b,n,c}\in\mathbb{R}$ is the feature value. The variance $\{\sigma^2\}_{l,c}=\frac{1}{B\times H_l\times W_l}\sum_b^B\sum_n^{H_l\times W_l} (f_{l,b,n,c}-\mu_{l,c})^2$. Then, the input feature is normalized as 
\begin{align}
     \tilde{f}_{l,b,n,c}=\gamma_{l,c}({f}_{l,b,n,c}-\mu_{l,c})/\sqrt{\{\sigma^2\}_{l,c}+\epsilon}+\beta_{l,c},
\end{align}
where $\epsilon\in\mathbb{R}^+$ is a small scalar for numerical stability. $\gamma_{l,c}$ and $\beta_{l,c}$ are learnable scaling and shifting parameters, respectively.

In testing, the input is usually a single sample rather than a batch with $B$ samples. Therefore, BN stores the exponentially weighted average of the batch statistics at the training stage and used it in testing. Specifically, the mean and variance over the training are tracked progressively, i.e.,
\begin{align}
\overline{\mu}_{l,c}^k=(1-\eta)\cdot\overline{\mu}_{l,c}^{k-1}+\eta\cdot{\mu}_{l,c}^k; ~~~~\{\overline{\sigma}^2\}^k_{l,c}=(1-\eta)\cdot\{\overline{\sigma}^2\}^{k-1}_{l,c}+\eta\cdot\{{\sigma}^2\}^k_{l,c},  
\end{align} where $\eta\in[0,1]$ is a momentum parameter. After $K$ training iterations, $\overline{\mu}_{l,c}^K$, $\{\overline{\sigma}^2\}^K_{l,c}$, $\gamma_{l,c}^K$, and $\beta_{l,c}^K$ are stored and used for testing normalization \cite{ioffe2015batch}. 
 
\subsection{Adaptive source-relaxed batch-wise statistics adaptation} 

Early attempts of BN for UDA simply added BN in the target domain, without the interaction with the source domain \cite{li2018adaptive}. Recent studies \cite{chang2019domain,maria2017autodial,wang2019transferable,mancini2018boosting} indicated that the low-order batch statistics, i.e., mean $\mu_{l,c}$ and variance $\{\sigma^2\}_{l,c}$, are domain-specific, because of the divergence of cross-domain representation distributions. Therefore, brute-forcing the same mean and variance across domains can lead to a loss of expressiveness \cite{zhang2020generalizable}. In contrast, after the low-order batch statistics discrepancy is partially reduced, with domain-specific mean and variance normalization, the high-order batch statistics, i.e., scaling and shifting parameters $\gamma_{l,c}$ and $\beta_{l,c}$, are shareable across domains \cite{maria2017autodial,wang2019transferable}. 

However, all of the aforementioned methods \cite{chang2019domain,maria2017autodial,zhang2020generalizable,wang2019transferable,mancini2018boosting} require the source data at the adaptation stage. To address this, in this work, we propose to mitigate the domain shift via the adaptive low-order batch statistics progression with momentum, and explicitly enforce the consistency of the high-order statistics in a source-relaxed manner.

\noindent\textbf{Low-order statistics progression with an exponential momentum decay scheme}. In order to gradually learn the target domain-specific mean and variance, we propose an exponential low-order batch statistics decay scheme. We initialize the mean and variance in the target domain with the tracked $\overline{\mu}_{l,c}^K$ and $\{\overline{\sigma}^2\}^K_{l,c}$ in the source domain, which is similar to applying a model with BN in testing \cite{ioffe2015batch}. Then, we progressively update the mean and variance in the $t$-th adaptation iteration in the target domain as
\begin{align}
\overline{\mu}_{l,c}^t=(1-\eta^t)\cdot{\mu}_{l,c}^{t}+\eta^t\cdot\overline{\mu}_{l,c}^K; ~~~~\{\overline{\sigma}^2\}^t_{l,c}=(1-\eta^t)\cdot\{{\sigma}^2\}^{t}_{l,c}+\eta^t\cdot\{\overline{\sigma}^2\}^K_{l,c},
\end{align} 
where $\eta^t=\eta^0\text{exp}(-t)$ is a target adaptation momentum parameter with an exponential decay w.r.t. the iteration $t$. ${\mu}_{l,c}^{t}$ and $\{{\sigma}^2\}^{t}_{l,c}$ are the mean and variance of the current target batch. Therefore, the weight of $\overline{\mu}_{l,c}^K$ and $\{\overline{\sigma}^2\}^K_{l,c}$ are smoothly decreased along with the target domain adaptation, while ${\mu}_{l,c}^{t}$ and $\{{\sigma}^2\}^{t}_{l,c}$ gradually represent the batch-wise low-order statistics of the target data.    

\noindent\textbf{Transferability adaptive high-order statistics consistency}. For the high-order batch statistics, i.e., the learned scaling and shifting parameters, we explicitly encourage its consistency between the two domains with the following high-order batch statistics (HBS) loss: 
\begin{align}
\mathcal{L}_{HBS}= \sum_l^L\sum_{c}^{C_l} (1+\alpha_{l,c}) \{|\gamma_{l,c}^K-\gamma_{l,c}^t| + |\beta_{l,c}^K-\beta_{l,c}^t|\},
\end{align} where $\gamma_{l,c}^K$ and $\beta_{l,c}^K$ are the learned scaling and shifting parameters in the last iteration of pre-training in the source domain. $\gamma_{l,c}^t$ and $\beta_{l,c}^t$ are the learned scaling and shifting parameters in the $t$-th adaptation iteration. $\alpha_{l,c}$ is an adaptive parameter to balance between the channels.

We note that the domain divergence can be different among different layers and channels, and the channels with smaller divergence can be more transferable \cite{pan2018two}. Accordingly, we would expect that the channels with higher transferability contribute more to the adaptation. In order to quantify the domain discrepancy in each channel, a possible solution is to measure the difference between batch statistics. In the source-relaxed UDA setting, we define the channel-wise source-target distance in the $t$-th adaptation iteration as
\begin{align}
    d_{l,c}=|\frac{\overline{\mu}_{l,c}^K}{\sqrt{\{\overline{\sigma}^2\}^K_{l,c}+\epsilon}}-\frac{{\mu}_{l,c}^t}{\sqrt{\{{\sigma}^2\}^t_{l,c}+\epsilon}}|.
\end{align} 
Then, the transferability of each channel can be measured by $\alpha_{l,c}=\frac{L\times C\times(1+d_{l,c})^{-1}}{\sum_l\sum_c(1+d_{l,c})^{-1}}$. Therefore, the more transferable channels will be assigned with higher importance, i.e., with larger weight $(1+\alpha_{l,c})$ in $\mathcal{L}_{l,c}$.

\subsection{Self-entropy minimization and overall training protocol}

The training in the unlabeled target domain can also be guided by an unsupervised learning framework. The SE minimization is a widely used objective in modern DNNs to encourage the confident prediction, i.e., the maximum softmax value can be high  \cite{grandvalet2005semi,liang2020we,wang2020fully,bateson2020source}. SE for pixel segmentation is calculated by the averaged entropy of the classifier's softmax prediction given by
\begin{align}
    \mathcal{L}_{SE}=
    \frac{1}{B\times H_0\times W_0}\sum_b^B\sum_n^{H_0\times W_0}\{{\delta_{b,n} \text{log} \delta_{b,n}}\},
\end{align}
where $H_0$ and $W_0$ are the height and width of the input, and $\delta_{b,n}$ is the histogram distribution of the softmax output of the $n$-th pixel of the $b$-th image in a batch. Minimizing $\mathcal{L}_{SE}$ leads to the output close to a one-hot distribution.

At the source-domain pre-training stage, we follow the standard segmentation network training protocol. At the target domain adaptation stage, the overall training objective can be formulated as $\mathcal{L}=\mathcal{L}_{HBS}+\lambda\mathcal{L}_{SE}$, where $\lambda$ is used to balance between the BN statistics matching and SE minimization. We note that a trivial solution of SE minimization is that all unlabeled target data could have the same one-hot encoding \cite{grandvalet2005semi}. Thus, to stabilize the training, we linearly change the hyper-parameter $\lambda$ from 10 to 0 in training.

\section{Experiments and Results}

\begin{savenotes}
\begin{table}[t!]
\caption{Comparison of HGG to LGG UDA with the four-channel input for our four-class segmentation, i.e., whole tumor, enhanced tumor, core tumor, and background. $\pm$ indicates standard deviation. SEAT \cite{shanis2019intramodality} with the source data for UDA training is regarded as an ``upper bound."} 
\label{tab1}
\centering
\resizebox{1\linewidth}{!}{
\begin{tabular}{c|c|cccc|cccc}
    \hline
    \multirow{2}*{Method}&Source& \multicolumn{4}{c|}{Dice Score [\%] $\uparrow$}& \multicolumn{4}{c}{Hausdorff Distance [mm] $\downarrow$} \\ \cline{3-10}
    &data& WholeT & EnhT & CoreT & Overall & WholeT & EnhT & CoreT & Overall \\ \hline \hline
     Source only & no UDA &79.29& 30.09& 44.11 &58.44$\pm$43.5 &38.7 &46.1 &40.2 &41.7$\pm$0.14\\\hline  
    CRUDA~\cite{bateson2020source} & Partial\footnote{An additional class ratio predictor was required to be trained with the source data.}& 79.85 &31.05& 43.92& 58.51$\pm$0.12 &31.7 &29.5 &30.2 &30.6$\pm$0.15\\\hline  
    \textbf{OSUDA} & \textbf{no} &\textbf{83.62} & \textbf{32.15}& \textbf{46.88} &\textbf{61.94$\pm$0.11} &\textbf{27.2} &\textbf{23.4} &\textbf{26.3} &\textbf{25.6$\pm$0.14}\\ 
           OSUDA-AC & no &82.74 & 32.04& 46.62 &60.75$\pm$0.14 &27.8 &25.5 &27.3 &26.5$\pm$0.16\\    
           OSUDA-SE & no &82.45 & 31.95& 46.59 &60.78$\pm$0.12 &27.8 &25.3 &27.1 &26.4$\pm$0.14\\     \hline      SEAT~\cite{shanis2019intramodality} & Yes& 84.11 & 32.67 & 47.11 & 62.17$\pm$0.15 &26.4 &21.7 &23.5 &23.8$\pm$0.16\\ \hline
\end{tabular}
}
\end{table}
\end{savenotes}

The BraTS2018 database is composed of a total of 285 subjects \cite{menze2014multimodal}, including 210 high-grade gliomas (HGG, i.e., glioblastoma) subjects, and 75 low-grade gliomas (LGG) subjects. Each subject has T1-weighted (T1), T1-contrast enhanced (T1ce), T2-weighted (T2), and T2 Fluid Attenuated Inversion Recovery (FLAIR) Magnetic Resonance Imaging (MRI) volumes with voxel-wise labels for the enhancing tumor (EnhT), the peritumoral edema (ED), and the necrotic and non-enhancing tumor core (CoreT). Usually, we denote the sum of EnhT, ED, and CoreT as the whole tumor. In order to demonstrate the effectiveness and generality of our OSUDA, we follow two UDA evaluation protocols using the BraTS2018 database, including HGG to LGG UDA \cite{shanis2019intramodality} and cross-modality (i.e., T2 to T1/T1ce/FLAIR) UDA \cite{zou2020unsupervised}.

For evaluation, we adopted the widely used Dice similarity coefficient and Hausdorff distance metrics as in \cite{zou2020unsupervised}. The Dice similarity coefficient (the higher, the better) measures the overlapping part between our prediction results and the ground truth. The Hausdorff distance (the lower, the better) is defined between two sets of points in the metric space. %The implementation details are given in \textbf{supplementary}. 

\begin{figure}[t]
\begin{center}
\includegraphics[width=1\linewidth]{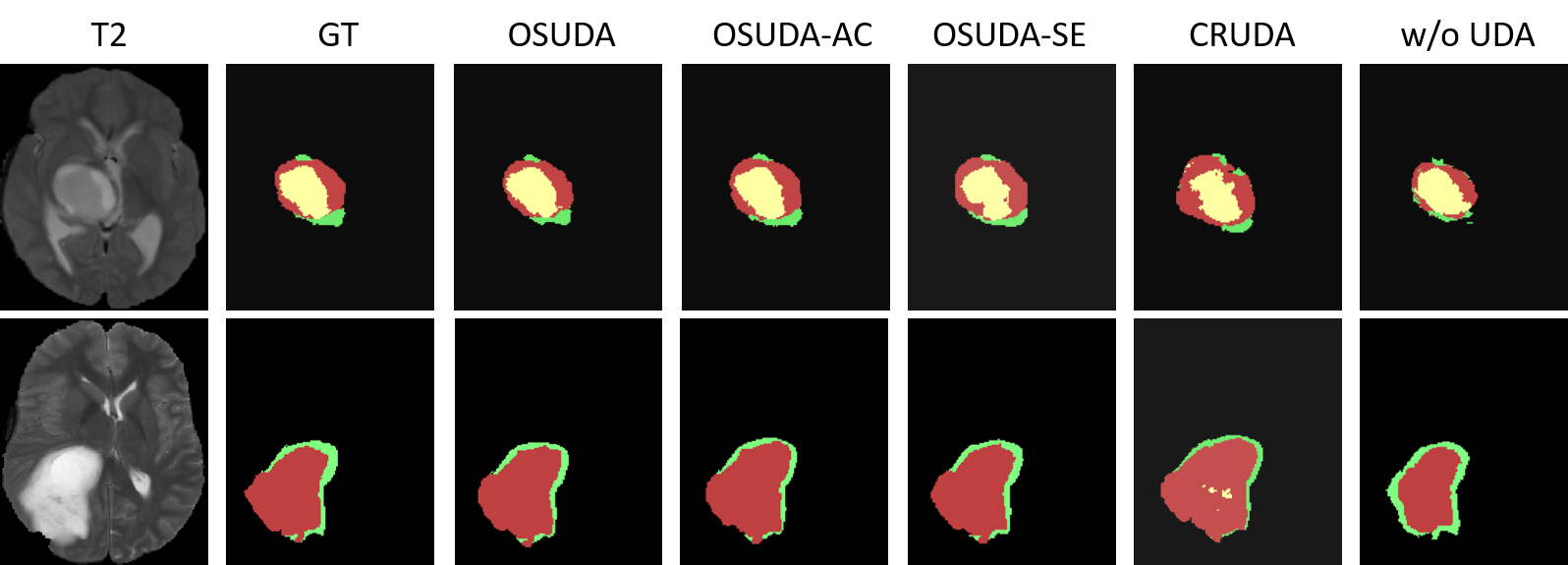}
\end{center}
\caption{The comparison with the other UDA methods, and an ablation study of adaptive channel-wise weighting and SE minimization for HGG to LGG UDA.} 
\label{exp1}
\end{figure}

\subsection{Cross-subtype HGG to LGG UDA}

HGG and LGG have different size and position distributions for tumor regions \cite{shanis2019intramodality}.
Following the standard protocol, we used the HGG training set (source domain) to pre-train the segmentation model and adapted it with the LGG training set (target domain) \cite{shanis2019intramodality}. The evaluation was implemented in the LGG testing set. We adopted the same 2D U-Net backbone in \cite{shanis2019intramodality}, sliced 3D volumes into 2D axial slices with the size of 128$\times$128, and concatenated all four MRI modalities to get a 4-channel input.

The quantitative evaluation results are shown in Table~\ref{tab1}. Since the pixel proportion of each class is different between HGG and LGG domains, the class ratio-based CRUDA \cite{bateson2020source} only achieved marginal improvements with its unsupervised learning objective. We note that the Dice score of the core tumor was worse than the pre-trained source-only model, which can be the case of negative transfer \cite{wang2019characterizing}. Our proposed OSUDA achieved the state-of-the-art performance for source-relaxed UDA segmentation, approaching the performance of SEAT~\cite{shanis2019intramodality} with the source data, which can be seen as an ``upper-bound."

We used OSUDA-AC and OSUDA-SE to indicate the OSUDA without the adaptive channel-wise weighting and self-entropy minimization, respectively. The better performance of OSUDA over OSUDA-AC and OSUDA-SE demonstrates the effectiveness of adaptive channel-wise weighting and self-entropy minimization. The illustration of the segmentation results is given in Fig. \ref{exp1}. We can see that the predictions of our proposed OSUDA are better than the no adaptation model. In addition, CRUDA \cite{bateson2020source} had a tendency to predict a larger area for the tumor; and the tumor core is often predicted for the slices without the core.

\subsection{Cross-modality T2 to T1/T1ce/FLAIR UDA}

\begin{table}[t]
\centering
\caption{Comparison of whole tumor segmentation for the cross-modality UDA. We used T2-weighted MRI as our source domain, and T1-weighted, FLAIR, and T1ce MRI as the unlabeled target domains.}
\resizebox{1\linewidth}{!}{
\begin{tabular}{c|c|cccc|cccc}
\hline

&Source& & Dice & Score & [$\%$] $\uparrow$ &  & Hausdorff & Distance & [mm] $\downarrow$ \\ \cline{3-10}

{Method}&data& T1 & FLAIR & T1CE & Average & T1 & FLAIR & T1CE & Average \\ \hline \hline

Source only & no UDA & 6.8 &54.4 &6.7& 22.6$\pm$0.17& 58.7& 21.5 &60.2 &46.8$\pm$0.15\\\hline

CRUDA~\cite{bateson2020source} & Partial\footnote{An additional class ratio predictor was required to be trained with the source data.}& 47.2 &65.6& 49.4& 54.1$\pm$0.16 &22.1 &17.5 &24.4 &21.3$\pm$0.10\\\hline

\textbf{OSUDA} & \textbf{no} & \textbf{52.7} &\textbf{67.6}& \textbf{53.2}& \textbf{57.8$\pm$0.15}& \textbf{20.4} &\textbf{16.6}& \textbf{22.8}& \textbf{19.9$\pm$0.08}\\

OSUDA-AC & no & 51.6 &66.5 &52.0& 56.7$\pm$0.16 &21.5 &17.8 &23.6 &21.0$\pm$0.12\\    

OSUDA-SE & no & 51.1 &65.8& 52.8& 56.6$\pm$0.14 &21.6 &17.3 &23.3 &20.7$\pm$0.10\\\hline \hline   
    
CycleGAN~\cite{zhu2017unpaired} & Yes&38.1 &63.3& 42.1& 47.8 &25.4 &17.2 &23.2 &21.9\\ 

SIFA~\cite{chen2019synergistic}& Yes& 51.7 &68 &58.2& 59.3 &19.6 &16.9& 15.01& 17.1\\

DSFN~\cite{zou2020unsupervised} & Yes& 57.3 &78.9 &62.2 &66.1& 17.5& 13.8 &15.5& 15.6\\ \hline
\end{tabular}}\label{tab2}
\end{table}

\begin{figure}[t!]
\begin{center}
\includegraphics[width=1\linewidth]{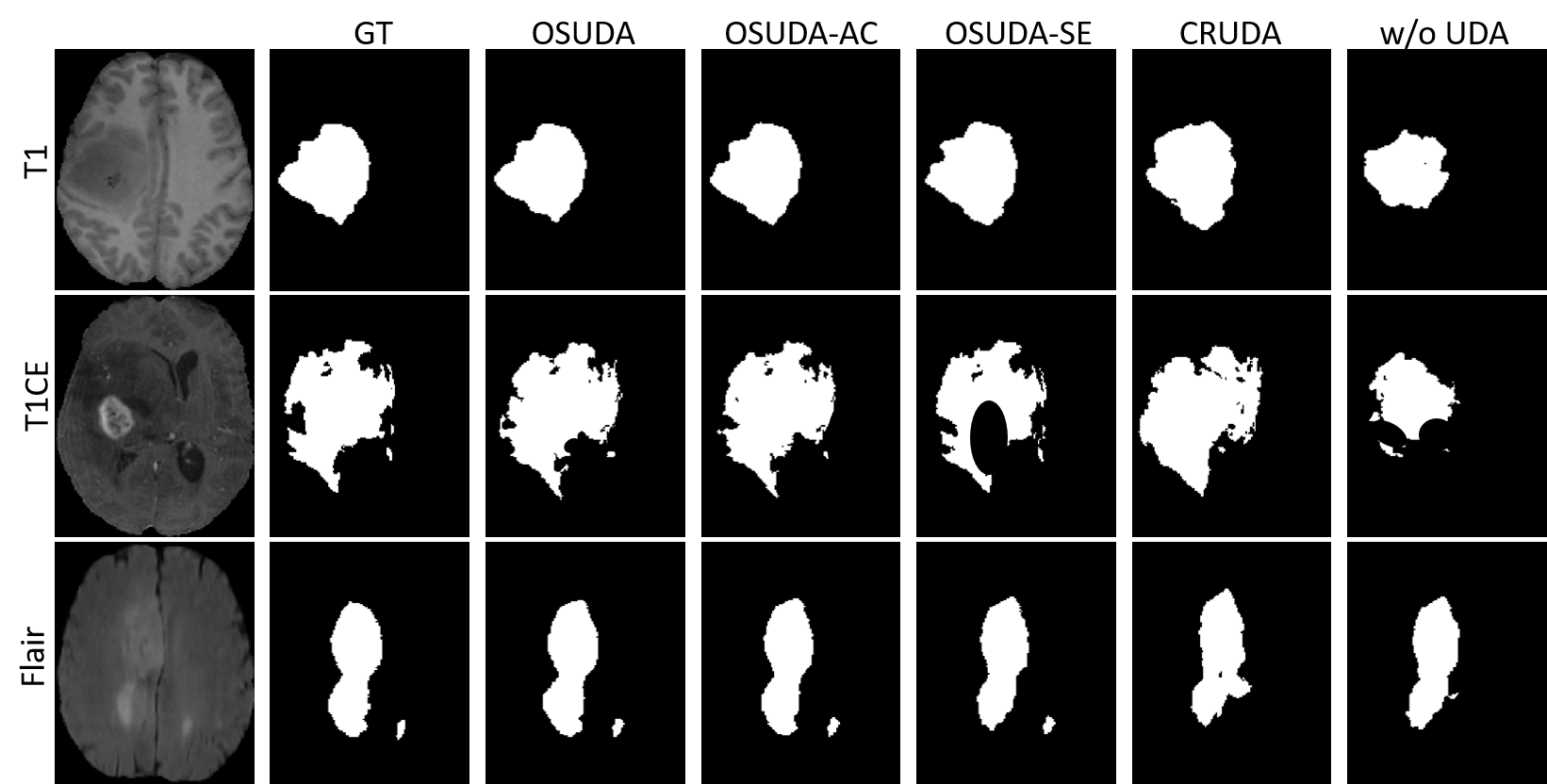}
\end{center} 
\caption{Comparison with the other UDA methods and an ablation study for the cross-modality whole tumor segmentation UDA task. From top to bottom, we show a target test slice of T1, T1ce, and FLAIR MRI.} 
\label{exp2}
\end{figure} 

Because of large appearance discrepancies between different MRI modalities, we further applied our framework to the cross-modality UDA task. Since clinical annotation of the whole tumor is typically performed on T2-weighted MRI, the typical cross-modality UDA setting is to use T2-weighted MRI as the labeled source domain, and T1/T1ce/FLAIR MRI as the unlabeled target domains \cite{zou2020unsupervised}. We followed the UDA training (80\% subjects) and testing (20\% subjects) split as in \cite{zou2020unsupervised}, and adopted the same single-channel input backbone. We note that the data were used in an unpaired manner \cite{zou2020unsupervised}.
 
The quantitative evaluation results are provided in Table~\ref{tab2}. Our proposed OSUDA outperformed CRUDA \cite{bateson2020source} consistently. In addition, in CRUDA, the additional class ratio prediction model was required to be trained with the source data, which is prohibitive in many real-world cases. Furthermore, our OSUDA outperformed several UDA methods trained with the source data, e.g., CycleGAN \cite{zhu2017unpaired} and SIFA \cite{chen2019synergistic}, for the two metrics. The visual segmentation results of three target modalities are shown in Fig. \ref{exp2}, showing the superior performance of our framework, compared with the comparison methods. 

\section{Discussion and Conclusion}

This work presented a practical UDA framework for the tumor segmentation task in the absence of the source domain data, only relying on the “off-the-shelf” pre-trained segmentation model with BN in the source domain. We proposed a low-order statistics progression with an exponential momentum decay scheme to gradually learn the target domain-specific mean and variance. The domain shareable high-order statistics consistency is enforced with our HBS loss, which is adaptively weighted based on the channel-wise transferability. The performance was further boosted with the unsupervised learning objective via self-entropy minimization. Our experimental results on the cross-subtype and cross-modality UDA tasks demonstrated that the proposed framework outperformed the comparison methods, and was robust to the class ratio shift.

\section*{Acknowledgements}

This work is partially supported by NIH R01DC018511, R01DE027989, and P41EB022544.

\bibliographystyle{splncs04}
\bibliography{egbib}

\end{document}